# Time-Aware Datasets are Adaptive Knowledgebases for the New Normal

Abhijit Suprem[1], Sanjyot Vaidya[1], Joao Eduardo Ferreira[2], and Calton Pu[1]

[1] School of Computer Science, Georgia Institute of Technology, USA
[2] Institute of Mathematics and Statistics, University of Sao Paulo, Brazil
`asuprem@gatech.edu`

**Abstract.** Recent advances in text classification and knowledge capture in language models have relied on availability of large-scale text datasets. However, language models are trained on static snapshots of knowledge and are limited when that knowledge evolves. This is especially critical for misinformation detection, where new types of misinformation continuously appear, replacing old campaigns. We propose time-aware misinformation datasets to capture time-critical phenomena. In this paper, we first present evidence of evolving misinformation and show that incorporating even simple time-awareness significantly improves classifier accuracy. Second, we present COVID-TAD, a large-scale COVID-19 misinformation dataset spanning 25 months. It is the first large-scale misinformation dataset that contains multiple snapshots of a datastream and is orders of magnitude bigger than related misinformation datasets. We describe the collection and labeling process, as well as preliminary experiments.

## 1 Introduction

**Timeless Knowledge.** Through the history of mankind, most changes happened relatively slowly, on the scale of years and decades. A significant part of human knowledge, e.g., laws of physics, have proven to be applicable over many years. This timeless knowledge assumption has been applied to artificial intelligence (AI) models. For example, it is only natural to assume that computer vision models trained to recognize apples and oranges would maintain their accuracy as long as the fruits continue to look the same way as they have looked historically. However, the world has been evolving in an increasing pace. The accelerating changes include long term trends such as technology advances and short-term shocks such as the COVID-19 pandemic. Specifically, the rapid mutations of SARS-COV-2 virus (e.g., subvariants of Omicron) have caused many, if not all, scientific and AI models to become quickly obsolete, often in a matter of weeks [1]. The key to this obsolescence is the expiration of facts: while some facts and knowledge are timeless, others have an expiration data. When ML classifiers incorporate timeless as well as temporally scoped knowledge together in their training data, they run the risk of obsolescence due to fact expiry.

Recent advances in large-language models [2] and foundation models [3] have significantly improved the state-of-the-art in image classification, generation, captioning, text classification, and text generation. LLMs and foundation models such as CLIP [4], GPT [5], and BERT [6] achieve these results because they are trained on significantly more data than prior counterparts (such as ResNet [7] and word2vec [8]). As such, recent works in classification, prediction, translation consider LLMs and foundation models as generalized knowledgebase stores capable of answering questions beyond the scope of their training data [9].

Such a claim is the gold-standard for machine learning pipelines in the wild, where a model is first pre-trained on a large knowledge base, then fine-tuned on a task-specific dataset. The fine-tuning effectively translates the pre-training knowledgebase into the task-specific domain. However, the pre-training knowledgebase contains facts and knowledge that are temporally scoped [10]: it contains only a snapshot of knowledge at the time of collection. Even with fine-tuning that adapts the training data embeddings of the classifier to the task-specific domain [11], relying on temporally scoped knowledge that may expire can have adverse detrimental impacts on long-term classifier performance [12].

**Knowledge Expiration.** There are several examples of classifier deterioration due to knowledge expiration: (i) Google Flu Trends [13], (ii) Uber self-driving accidents [14], and (iii) Microsoft Tay Chatbot [15]. Google Flu Trends (GFT) [16] was designed to predict the peak of the flu based on search trends. It was initially released in 2008, where it achieved over 90% accuracy in predicting flu trends. In subsequent years, due to fixed initial-training and a static update procedure that neglected new search terms and lexical drift [13], GFT made significant errors, missing the flu season with greater than 100% errors (i.e. completely missing flu peaks) [17] . Uber's self-driving cars, initially trained in Arizona, caused fatalities when tested in a non-desert environment due to failures of domain shift approaches for unknown test distributions. The self-driving vehicles could not distinguish white trucks from the sky, or pedestrians outside of crosswalks [14]. Finally, Microsoft Tay was an AI chatbot trained to learn from its conversations. However, bad-faith actors exploited lack of safety checks to add inappropriate phrases and slurs into Tay's vocabulary [15]. More recently, the impact of using fixed-dataset temporally scoped LLMs is explored in [18]: state-of-the-art LLMs without such time-awareness face significant accuracy decline when tested on a diagnostic dataset of temporally scoped facts from multiple windows (e.g. questions about who is governor of a state, or which team LeBron James plays for in 2015 and 2020) [10].

As such, a model trained on a snapshot of knowledge relies on stationarity of the snapshot for its predictions. When there is change in the real data stream, such as in the marginal $P(X)$ as well as the conditional $P(Y|X)$, the fixed-window approach results in detrimental classifiers due to the assumption that facts never expire. We call this the timeless knowledge assumption. It is important to note that only a subset of facts expire, and that the timeless knowledge assumption is valid if the task relies on such forever facts (e.g. age of the universe, shape of a ball) [19].

**Knowledge Expiration in Misinformation.** Let us consider misinformation and fake news detection. Modern social media have facilitated the spread of misinformation on evolving topics such as COVID infodemic [20]. With impact on mission-critical and life-threatening topics such as vaccine hesitancy, research on accurate and timely detection of misinformation and fake news is urgently needed. As examples of accurate detection, state-of-the-art approaches based on expert systems built around pre-trained language models (PLMs) [21] with refinements (e.g., social context, and keyword attention net-works) have been trained on curated, fixed datasets and tested on data matching the training distribution. Unfortunately, as the infodemic evolves with the pandemic, these state-of-the-art approaches have increasing difficulties as time passes, as we discuss in Evaluation.

In retrospect, the decreasing performance of machine learning (ML) models trained from fixed ground truth should not be surprising, since new pieces (fake news disseminated after the training data) are completely unknown to those ML models. For cases where models are expected to operate on streaming real-world data, a time-aware knowledge assumption supersedes a timeless knowledge assumption. Time-aware knowledge assumption considers the possibility of new knowledge adding to existing knowledge or replacing existing knowledge.

To fully encapsulate time-aware knowledge, we introduce the concept of new-normal, a term which is also used to describe the post-COVID scene of emerging variants and rapidly changing institutional knowledge and policy. We first present it in context of COVID misinformation, then extend it to real-world streams.

**New-Normal.** We borrow the term "new normal" from the COVID-19 pandemic [22] to describe the challenges introduced by accelerating changes. Similar to the evolution of coronavirus variants, the new-normal data streams have two important properties that challenge the timeless knowledge assumption in classic ML training data sets. Unlike the gold standard fixed, curated, and timeless ground truth, the growing new-normal data streams contain: (1) never-seen-before novelty, and (2) short-lived abundance. By never-seen-before novelty, we mean natural or man-made truly novel creations such as the Omicron variant or the 5G theory of the origin of the SARS-COV-2 virus. By short-lived abundance we refer to the immediate prevalence of new-normal, typically followed by a decline, often due to the next new-normal. In the genetic evolution of SARS-COV-2 variants, the succession of temporarily dominant variant (starting from the Alpha and continuing with the sub-variants of the Omicron) exemplify short-lived abundance. Similarly, the succession of conspiracy theories for the origins of COVID-19 pandemic also illustrate the short-lived abundance of new-normal.

**Challenges of New-Normal.** The two properties of new-normal cause significant difficulties to the gold standard fixed, curated ground truth. First, consider the theoretically perfect model (from a fixed, curated training data) that achieves zero false positives and zero false negatives. By its very nature, the never-seen-before novelty cannot be present in any curated training data prior to the creation of such novelty. There-fore, even hypothetical perfect model would be ignorant of never-seen-before novelty, with unpredictable accuracy from knowledge that was perfect, but now obsolete. Second, the new-normal cannot be ignored due to the impact of its abundance when they arise, exacerbating the knowledge obsolescence problem. Third, the short-life of each wave leads to continual knowledge obsolescence, as new waves of new-normal relentlessly replace the previous ones.

**Urgency of New-Normal.** The new-normal has emerged from the COVID-19 pandemic, and it has become a growing part of our reality [22], changing our world and lives. It is important and urgent to take the knowledge obsolescence challenges of new-normal by capturing it, studying it, and incorporating it into our ML models as each wave arrives. In this paper, we describe our efforts to collect continuously the new-normal of COVID-19 fake news. With significant influence on vaccine hesitancy, fake news has exerted and continues to exert significant impact on our society and human lives. The COVID-19 fake news exhibit fully the two properties of new-normal: first, the fake part of the COVID-19 fake news is never-seen-before novelty, e.g., wildly imaginative conspiracy theories such as the 5G cause of the pandemic. Second, the fake news are created and injected into social media en masse (abundance). Third, as their novelty wear off and credibility decline, the previous waves of fake news are replaced by new campaigns, starting another cycle of the new-normal.

**Need for Time Awareness.** Following the gold standard ML practice of studying fixed, curated data sets, it would be acceptable to wait for the rise and fall of each new-normal wave to stabilize, and then incorporate them into new models by curating the stabilized data. However, the incorporation of new data without timestamps introduce misinterpretations [23] when a clear trend (increasing or decreasing) is obscured by an average number. In order to capture and understand time-critical phenomena such as the rise and fall of new-normal, time awareness is essential. The need for time aware ML models can be observed as far back as the gradual decline of Google Flu Trends [13, 16, 17], and recently in Temporal Knowledge Bases [10, 24]. To enable the construction and evaluation of time-aware ML models, we need time aware data sets. This can be extended to time-aware pipelines, which include time-aware datasets as well as time-aware classifiers. In this work, we focus on time-aware datasets.

**Time-Aware Datasets.** In this work, we propose applying time-awareness to the training data as well as to the classifiers, thereby ensuring pipeline time-awareness. We can do this by using Time-Aware Datasets (TAD), which incorporate training data update and subsequent knowledge propagation to classifier weights directly into the training pipeline. Essentially, we convert the timeless knowledge approach of a single pre-trained knowledgebase to an evolving set of knowledgebases that are periodically refreshed. These updated knowledgebases, which snapshot a specific time window during their collection, subsequently generate new pre-trained bases for fine-tuning on their temporal scope. Since ML models are trained from ground truth data sets, the research and development of time-aware classifiers relies on such time-aware datasets, where each data item has a timestamp.

TADs are organized as a sequence of timed subsets, each pertaining to a well-defined time period (window). Each period should have sufficient a number of samples (not too narrow, to be representative and less vulnerable to small sample effect) that are clustered closely. Assuming abundance in new-normal, the appropriate time period is defined by the speed of change: faster changes indicate shorter periods for generating new windows.

**Contributions.** Our contributions are as follows:
1. COVID-19 Misinformation Time-Aware Dataset. We present a large-scale COVID-19 misinformation dataset spanning 25 months. It is the first large-scale misinformation dataset that contains multiple snapshots of a datastream.
2. Evidence of New-Normal. With our COVID-TAD dataset, we demonstrate evidence of knowledge obsolescence for misinformation detectors. We compare fixed-dataset, fixed-window, and time-aware classifiers on fake news detection capabilities, and show that time-aware classifiers significantly outperform their fixed-dataset and fixed-window counterparts.

## 2 Related Work

### 2.1 Relative Novelty

Concept drift [25] is an area focused on the detection of evolutionary changes of statistical properties in a data stream. Significant advances have been made in change detection [26] and drift adaptation [27]. These algorithms operate under specific assumptions. For example, virtual concept drift algorithms [28] are dedicated to detecting limited changes within fixed data sets. In contrast, the new-normal are more closely related to real concept drift, where absolute novelty (non-stationary new data) arrive continuously and unboundedly.

Evaluation experiments have been conducted primarily with limited data sets or under limiting assumptions such as fixed boundaries that can be modeled by stationary models. For example, continual learning [19] has focused on the prevention of, and recovery from, catastrophic forgetting [18] under timeless knowledge assumptions. In a slowly changing environment, absolute novelty appears rarely. For example, algorithms for anomaly detection [29], often assume the anomalies to be rare and known events.

The new-normal differ from the classic outlier and anomaly detection algorithms in two aspects. First, the absolute novelty (unknown unknowns) fall outside of previous knowledge, effectively expanding the data space beyond previous models. Second, the new-normal become dominant after their appearance, violating the rarity assumption. In practice, e.g., MLOps with tools such as Watson OpenScale [30], adaptation to absolute novelty (including new-normal) is expected to require new training data and resulting new models, e.g., through active learning [31].

## 2.2 Impermanence of New-Normal

If timeless knowledge assumption were applicable, retraining through active learning [31], dynamic model updates [18], and model confidence [32] would raise the performance of the accumulated retrained model. However, the new-normal typically arrive in multiple, unrelenting waves, overlapping and superseding each other, leading to accuracy decline when relying only on model updates and confidence without expanding the knowledgebase [12, 18, 24]. To the best of our knowledge, there are few papers (and data sets) that discuss or study the impermanence of data sets, since fixed data sets with implicit timeless knowledge assumption have been the gold standard in ML research.

## 2.3 Prior Research in Fake News

Fake news have been studied extensively through social science methods and ML analytics [12, 21, 24, 33-37]. The many papers on fake news re-search have confirmed their significant impact on humans and society. They also have demonstrated the excellent effectiveness of ML models in identifying fake news when trained from, and tested on, annotated data sets (e.g., k-fold validation). However, it is well known that they suffer from difficulties in generalizability [12], which has been attributed to data quality [20, 38, 39]. In case of fake news, it should not be surprising that an ML model trained on 2016 election fake news [12] would encounter difficulties when faced with (new and completely unrelated) fake news in 2022.

# 3 COVID-TAD

We now present our COVID-19 Fake News Time-Aware Dataset (COVID-TAD). First, we cover our data collection and labeling approach. Then we discuss our pre-training approach to generate time-aware pre-training models for subsequent fine-tuning.

## 3.1 Data Collection

We primarily focused on Twitter posts, and used links on tweets to extract articles, images, and videos. This allows us a variety of sources, such as short text social media, news articles, captioned images (with alt text), uncaptioned images, as well as links to videos posted anywhere and linked on Twitter. Our collection endpoint is a Twitter Sampled Stream, delivering a ~1% sample of all tweets. We apply keyword filters that are updated frequently during TAD generation across windows. When filters are updated, we reapply them to earlier windows to include any potentially missed early fake or real information.

Table 1. Keywords used for initial data collection

| Keyword | Related variations | | |
|---|---|---|---|
| "covid" | "corona" | "sars" | "pandemic" |
| "omicron" | "delta" | | |
| "mask" | "n95" | "ppe" | |
| "wuhan" | "lock" | | |
| "quarantin" | "social" | "travel" | |
| "virus" | "infect" | "variant" | |
| "vaccine" | "fauci" | | |
| "ivermectin" | | | |
| "plandemic" | "hoax" | "bioweapon" | |
| "5g" | "gates" | | |
| "bat" | "monkey" | | |

We use the minimal possible keyword for filtering, so "COVID", "COVID-19", and "COVID19", along with capitalization variations are all captured with the `covid` keyword. **Table 1** provides a list of base keywords, plus variations for each keyword. Due to potential bias towards English tweets, we also used Spanish and French translations of the keywords. Our dataset spans over 24 months, starting from January 25, 2020 and continuing till date. During this time, we have collected over 2TB worth of data, comprising over 2B tweets.

As a TAD, COVID-TAD has several important distinctions from traditional datasets: it contains multiple windows, and each window contains multiple subparts to narrow the scope of pre-training for each window. We generate 2 types of windows: fixed-window and adaptive windows. For fixed window, we generate monthly partitions of the full stream. For adaptive windows, we calculate relevance of each day's tweets with the prior day; when the relevance value, obtained from point-proximity overlap [ref], falls below a threshold, we trigger a new window.

We denote subcomponents of COVID-TAD as follows:

- o **COVID-TAD-Raw**: The raw stream collected from the Twitter Sampled Stream
- o **COVID-TAD-Extended**: The Raw stream is annotated with labels from multiple fake news detectors and fact checkers. We refer to monthly windows as COVID-TAD-[MonthYear], e.g. COVID-TAD-Jan2020
- o **COVID-TAD-Oracle**: We choose a subset of each COVID-TAD-Extended window for human labeling. The subset is selected such that it represents the distribution of the window. We aim for 400 samples per monthly window; over 25 months, this gives us 10k oracle samples.

### 3.2 Labeling Approach

We now describe our labeling approach for COVID-TAD-Extended and COVID-TAD-Oracle. First, we will describe how we selected subsets from the sampled stream for labeling in case of COVID-TAD-Extended and COVID-TAD-Oracle.

Then we will cover the actual labeling process. We will refer to the original sampled stream without keyword filtering as COVID-TAD-Raw.

**Generating COVID-TAD-Extended.** As we have mentioned, keyword filtering identifies the most relevant tweets to COVID-19, whether they are misinformation or not. However, it cannot account for variations in keywords, such as misspellings, obfuscation, or alternate spellings. For example, `quarantine` related posts also use the term `lockdown`, `isolation`, `sick`, and any variation of these words. Exhaustive keyword enumeration and search are impractical.

Here, we can exploit a property of LLMs like BERT and its variations. The pre-trained BERT comprises of an encoder and a decoder, each consisting of 6 self-attention transformer blocks. During pre-training, the encoder is provided text to encode into a 768-dimensional feature space. Subsequently, the decoder reconstructs the original text from the feature space representation. The encoder ensures semantically similar sentences occur close together in the feature space. This process, known as disentanglement, is crucial to extracting relevant context from text for classification.

We can apply this property to identify posts in COVID-TAD-Raw that were not selected by our keyword filter due to misspellings or obfuscation (e.g. by using alternate Unicode characters to fool keyword filters [40, 41]). The steps are:

1. **Fine-tune LLM.** Fine-tune a pretrained LLM, such as BERT using semantic masking on the keyword filtered data from COVID-TAD-Raw
2. **Generate Features.** Use the fine-tuned LLM to generate features for unfiltered data from COVID-TAD-Raw
3. **High density sets.** Use high density sets of the finetuned-LLM training data to identify samples in unfiltered data that are close to filtered data
4. **Generate COVID-TAD-Extended.** Generate COVID-TAD-extended by combining filtered data with nearby points in unfiltered data

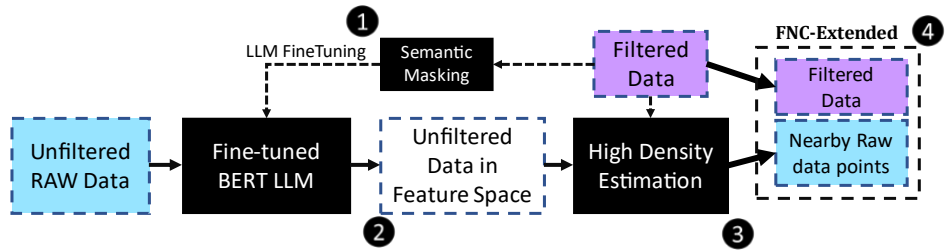

*Fine-tuning LLM.* First fine-tune a pre-trained LLM, such as BERT, on one window of the keyword filtered data. In our case, a window is a day's worth of data. A key challenge is the strong signal from keywords: a fine-tuned model has over-reliance on a small set of keywords it identifies within the training data [ref]. In our case, since the training data is exclusively obtained with keyword filtering, a fine-tuned LLM will identify the keywords themselves as representative of context. This problem is exacerbated when some keywords are strongly tied to the true class label; for example, the `plandemic` keyword almost exclusively appears with misinformation.

We can address this keyword strong signal using semantic masking. Commonly, LLM pre-training and fine-tuning employs a masked language modeling (MLM) transformation. With classic MLM proposed in [6], a fraction of input tokens are randomly masked to improve reconstruction generalization accuracy. Whole-word masking, proposed in [42] is an improvement on the token masking as it masks specific words, instead of individual tokens that might carry little semantic meaning. We apply a static mask to keywords from **Table 1**. With masking, we can expand our coverage on COVID-TAD-Raw. On average, we increase the size of the filtered subset by 23%, with a lift of 35% from the unfiltered dataset.

| Window | Extension | Lift |
|---|---|---|
| **January** | 4.18% | 11.70% |
| **February** | 38.38% | 58.90% |
| **March** | 7.40% | 18.59% |
| **April** | 39.90% | 60.88% |

*Generate Features.* We then use our fine-tuned LLM from each window to generate features for the unfiltered subset of COVID-TAD-Raw. When applied to the unfiltered subset, we can then find samples that are semantically similar to the training data of the LLM (i.e. filtered subset of COVID-TAD-Raw) by measuring distance in the feature space between unfiltered and filtered samples. This requires both a distance metric and a distance threshold. We accomplish both by computing the high-density set on the unmasked training features.

*High-Density Sets.* Neural networks, including transformers, perform disentanglement to cluster semantically similar samples in their lower dimensional feature space. This allows for easier classification from the feature space. First, we identify clusters in the feature space of the filtered subset. We can empirically determine the number of clusters with the elbow metric on a sweep of different values of $k$ in KMeans clustering.

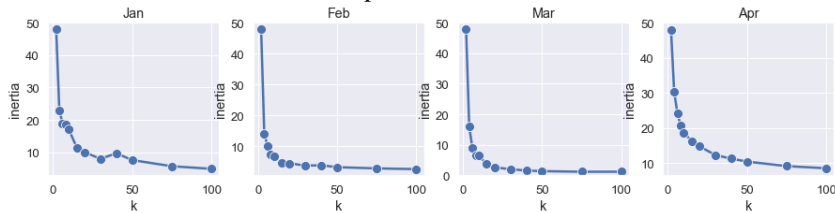

**Fig. 1.** Selecting optimal clusters for each window using KMeans sweep for different cluster values, followed by ELBOW metric

We show in **Fig. 1** examples of applying the elbow metric with cluster inertia for several windows of data. In each of the above cases, we select $k = 20$ as the optimal number of clusters, with the exception of April 2020, where we use $k = 50$.

Once we have an optimal number of clusters, we apply KMeans to generate cluster centers in the feature space of the filtered data. In this case, we use the unmasked versions of the filtered data, since we want to measure similarity between filtered and unfiltered samples. Once we have clusters of the filtered subset, we need to identify the high-density threshold of each cluster. The high-density threshold is adapted from level-set estimation and TrustScore in [32]; it is the radius within which the majority of samples exist for each cluster. To find the radius threshold for the high-density set of each cluster we use the radius with the majority (>50%) of samples in each cluster.

**Generate COVID-TAD-Extended.** Now, we use the fine-tuned LLM to project unfiltered samples into the feature space. Using a KDTree on cluster centers of the training data, we find the subset of unfiltered samples that exist inside any high-density set. This subset of unfiltered samples is combined with the filter subset to yield COVID-TAD-Extended. On average, we can increase the size of our dataset by including nearby unfiltered samples by ~25%. This corresponds to an average lift of 60% on the unfiltered samples themselves (**Fig. 2**).

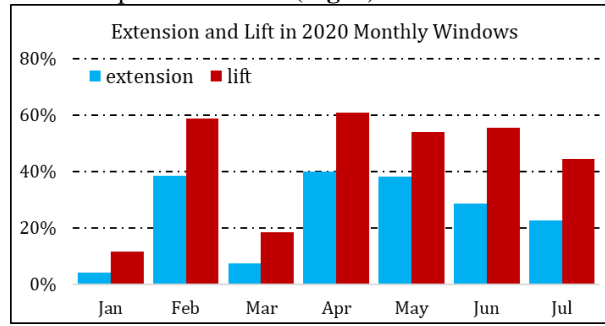

**Fig. 2.** Extension and Lift from the keyword filtered subset due to masking

**Generate COVID-TAD-Oracle.** We use COVID-TAD-Extended for our weak supervision labeling with multiple experts. To evaluate our labeling approach and to provide a human-labeled gold ground truth, we also generate COVID-TAD-Oracle. COVID-TAD-Oracle is a representative subset of COVID-TAD-Extended that is labeled with manual annotations. To generate COVID-TAD-Oracle, we selected 400 samples for each window for manual labeling. We first cluster each window with its optimal number of clusters, then bin samples (with 400 bins) based on distance to the cluster center using a KDTree. Finally, we select one sample per bin per cluster yielding 400 samples for each window evenly distributed across COVID-TAD-Extended

*Annotation Procedure.* We use multiple human annotators. Each sample is labeled by five annotators. Annotators are provided the text sample, username, likes, shares, retweets, and whether the content was subsequently deleted by the social media site. We accept labels that have agreement from all five annotators. Due to potential disagreements, we generated 500 samples for each window, and used the 400 samples with full agreement amongst annotators with random sampling. Further, with 500 samples, we have flexibility to ensure a class balanced labeled set.

**Labeling COVID-TAD-Extended.** The multi-source weak supervision labeling paradigm uses multiple experts or labeling functions to provide noisy labels for a prediction sample. These noisy labels can be aggregated to estimate the true latent labels. We aggregate labels by combining the PGM of Snorkel [43], neural network of WeaSEL [44], and smoothness calculations from MiDAS [45].

*Multiple Experts.* To ensure scalability, we use a team of experts to label COVID-TAD-Extended. Experts are deep learners trained on fake news datasets similar to COVID-TAD-Extended. We provide a list of datasets and statistics in **Table 2**. Each dataset contains COVID-19 related misinformation and true information. We train a BERT model on each dataset with masked language modeling and cross-entropy losses. Once experts are trained, we generate labels for each COVID-TAD-Extended window with the experts.

### 3.3 COVID-TAD Dataset Release

We provide a subset of COVID-TAD Tweet IDs[1]. Due to Twitter terms of service, we cannot publicly release the entire hydrated dataset, but it may be obtained upon request. The dataset contains multiple windows of tweets, totaling over 1B samples.

## 4 Evaluation of COVID-TAD

We describe our validation of knowledge expiration and the time-aware assumption. We first validate the need for time-aware datasets by exploring knowledge overlap and expiration in public fake news datasets. We use 11 fake news datasets discussed in [12]. We then explore the time-aware knowledge assumption in our COVID-TAD dataset by examining knowledge expiration between windows.

For each experiment, we train a variety of text classifiers:
1. BERT variants: We train text classifiers based on BERT, Covid-Twitter-BERT, and AlBERT. We use only the classifier head with the pre-trained base to fine-tune the feature extractor encoder.
2. SocialContext: Using the approach from NELA [46] and Fakeddit [47], we implement a SocialContext classifier using the text content encoded with BERT, plus share counts, like counts, and sentiment value.
3. MultiInput: We adapt the multi-domain approach from MDAWS [21]: we use a SocialContext branch weakly supervised by three labeling functions: swear word count, second-person pronoun count, and adverb count.

---

[1] https://grait-dm.gatech.edu/edna-covid-tweets-dataset/

### 4.1 Evaluations

*Knowledge expiration on public datasets.* We use public fake news datasets from [12], shown in **Table 2**, to validate the time-aware knowledge assumption. Given $n$ related datasets, e.g, fake news, datasets, we train classifiers on 1 fake news dataset and test on remaining $n-1$ datasets Since all datasets are fake news, a cross-domain classifier should perform well We perform some preliminary experiments using evaluation classifiers in each dataset: (i) same-dataset accuracy shows the test accuracy of a classifier trained on the same dataset; (ii) cross-dataset accuracy shows test accuracy of a classifier on remaining non-training datasets, and (iii) similar-dataset accuracy shows accuracy of classifiers when tested on datasets with similar content.

**Table 2.** Public Fake News datasets and baseline accuracies.

| Dataset | Train-Test Splits | Type | Same Dataset Acc. | Cross-Dataset Acc. (CDA) | Similar-Dataset Acc. (SDA) |
|---|---|---|---|---|---|
| k_title | 31K/9K | Titles | 0.97 | 0.50 | 0.60 |
| coaid | 5K/1K | Articles | 0.97 | 0.60 | 0.63 |
| c19_text | 2.5K/0.5K | Articles | 0.98 | 0.73 | 0.61 |
| cq | 12.5K/2K | Tweets | 0.54 | 0.53 | 0.51 |
| miscov | 4K/0.6K | Titles | 0.55 | 0.49 | 0.50 |
| k_text | 31k/9K | Articles | 0.98 | 0.56 | 0.57 |
| rumor | 4.5K/1K | Tweets | 0.83 | 0.52 | 0.55 |
| cov_fn | 4K/2K | Tweets | 0.96 | 0.46 | 0.51 |
| fakeddit | 878K/92K | Tweets | 0.80 | 0.49 | 0.64 |
| nela | 699K/158K | Tweets | 0.72 | 0.53 | 0.60 |
| c19_title | 2.5K/0.5K | Titles | 0.95 | 0.49 | 0.63 |

For example i.e. we test the 'k title' classifier on 'c19 title' and 'miscov' since they are both datasets of fake news titles. We have shown averaged results across classifiers, since performance was similar in each case. Due to knowledge expiration between training and testing data snapshots in cross-dataset setting, classifiers have lower accuracy compared to same-dataset setting.

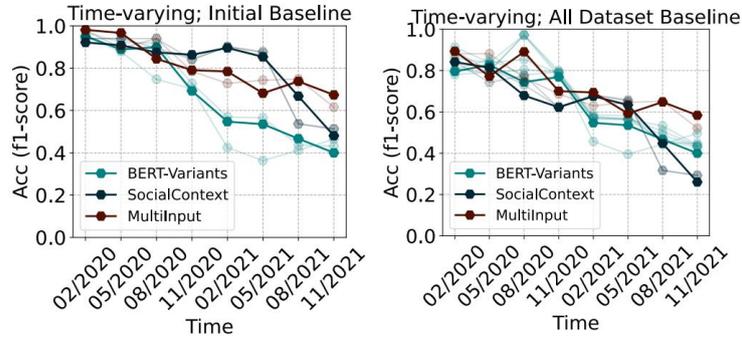

**Fig. 3.** Time-dependent experiment on COVID-TAD. For graph on left, training data is first three windows of COVID-TAD. For graph on right, training data is first 3 windows of COVID-TAD plus all datasets from **Table 2**

*Knowledge expiration on COVID-TAD.* Now we cover knowledge expiration on COVID-TAD with a time-dependent experiment. As a starting point, we trained BERT, SocialContext, and MultiInput classifiers on the first 3 months of COVID-TAD-Extended windows. We then tested them on each COVID-TAD-Extended window. We show representative experiments in **Fig. 3** with remaining experiments as faded lines. Accuracy declines on subsequent windows in COVID-TAD, similar to accuracy decline in **Table 2**, due to knowledge expiration.

To better gauge the impact of knowledge expiration, we modified the experiment to include public fake news data as well: that is, we trained BERT, SocialContext, and MultiInput classifiers on the first 3 month windows of COVID-TAD-Extended, as well as on the 11 datasets in **Table 2**. We show results in the right side graph in **Fig. 3**.. These performed slightly worse due to distributional differences in the training sets. The BERT variants show significant performance degradation, leading towards random guess accuracy for most models. SocialContext and MultiInput approaches have slower performance degradation due to additional knowledge used in misinformation prediction. Over time, they also approach random guess accuracy.

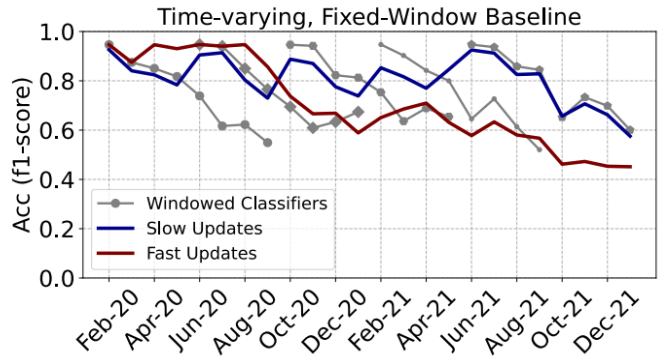

**Fig. 4.** Time-dependent experiment with time-aware classifier replacement

*Time-Varying, Windowed-Baseline.* Finally, we apply a simple update scheme, where for each window, we use the masking LLM from the COVID-TAD-Extended generation step of the prior window as a pre-training base. We then fine-tune the LLM on the labels for the prior window, and use it to predict labels for the current window. This effectively updates our classifier in each window with new knowledge. We compare 2 update settings: (i) slow updates, where we refresh our classifier with bimonthly windows, and (ii) fast updates, where we refresh our classifier with monthly windows, shown in **Fig. 4**. Fast updates led to catastrophic forgetting, where models approached random-guess after a few updates due to replacement of important weights inside the classifiers (red in **Fig. 4**). Slower updates lead to less accuracy decline.

## 5    Conclusions

Most knowledge is temporally scoped. We showed extensive examples with respect to misinformation. However, there are few comparable datasets with temporal scoping and as such, little exploration on the impact of time-awareness on knowledge capture and modeling. In this paper, we presented evidence of knowledge obsolescence and the need for time-awareness in misinformation detection. We have also presented COVID-TAD, a large-scale, temporally scoped, time-aware dataset for misinformation detection spanning over 25 months. COVID-TAD contains over 1 billion social media posts relevant to COVID, with annotations from multiple labelers. We described COVID-TAD's collection and labeling process, and release COVID-TAD for public use.

## Acknowledgements

This research has been partially funded by National Science Foundation by CISE/CNS (1550379, 2026945, 2039653), SaTC (1564097), SBE/HNDS (2024320) programs, and gifts, grants, or contracts from Fujitsu, HP, and Georgia Tech Foundation through the John P. Imlay, Jr. Chair endowment. Any opinions, findings, and conclusions or recommendations expressed in this material are those of the author(s) and do not necessarily reflect the views of the National Science Foundation or other funding agencies and companies mentioned above.

## References

[1]    N. G. Davies *et al.*, "Estimated transmissibility and impact of SARS-CoV-2 lineage B.1.1.7 in England," *Science,* vol. 372, no. 6538, p. eabg3055, 2021, doi: doi:10.1126/science.abg3055.

[2]    K. Sun, X. Luo, and M. Y. Luo, "A Survey of Pretrained Language Models," Springer International Publishing, 2022, pp. 442-456.

[3]    R. Bommasani *et al.*, "On the opportunities and risks of foundation models," *arXiv preprint arXiv:2108.07258,* 2021.

[4]     A. Khandelwal, L. Weihs, R. Mottaghi, and A. Kembhavi, "Simple but effective: Clip embeddings for embodied ai," in *Proceedings of the IEEE/CVF Conference on Computer Vision and Pattern Recognition*, 2022, pp. 14829-14838.

[5]     K. Ethayarajh, "How contextual are contextualized word representations? comparing the geometry of BERT, ELMo, and GPT-2 embeddings," *arXiv preprint arXiv:1909.00512,* 2019.

[6]     J. Devlin, M.-W. Chang, K. Lee, and K. Toutanova, "Bert: Pre-training of deep bidirectional transformers for language understanding," *arXiv preprint arXiv:1810.04805,* 2018.

[7]     K. He, X. Zhang, S. Ren, and J. Sun, "Deep residual learning for image recognition," in *Proceedings of the IEEE conference on computer vision and pattern recognition*, 2016, pp. 770-778.

[8]     T. Mikolov, E. Grave, P. Bojanowski, C. Puhrsch, and A. Joulin, "Advances in pre-training distributed word representations," *arXiv preprint arXiv:1712.09405,* 2017.

[9]     F. Petroni *et al.*, "Language models as knowledge bases?," *arXiv preprint arXiv:1909.01066,* 2019.

[10]    B. Dhingra, J. R. Cole, J. M. Eisenschlos, D. Gillick, J. Eisenstein, and W. W. Cohen, "Time-Aware Language Models as Temporal Knowledge Bases," *Transactions of the Association for Computational Linguistics,* vol. 10, pp. 257-273, 2022, doi: 10.1162/tacl_a_00459.

[11]    M. F. Chen *et al.*, "Shoring Up the Foundations: Fusing Model Embeddings and Weak Supervision," *arXiv preprint arXiv:2203.13270,* 2022.

[12]    A. Suprem and C. Pu, "Evaluating Generalizability of Fine-Tuned Models for Fake News Detection," *arXiv preprint arXiv:2205.07154,* 2022.

[13]    L. Kugler, "What happens when big data blunders?," *Communications of the ACM,* vol. 59, no. 6, pp. 15-16, 2016.

[14]    S. He, "Who is liable for the UBER self-driving crash? Analysis of the liability allocation and the regulatory model for autonomous vehicles," in *Autonomous Vehicles*: Springer, 2021, pp. 93-111.

[15]    T. Zemčík, "Failure of chatbot Tay was evil, ugliness and uselessness in its nature or do we judge it through cognitive shortcuts and biases?," *AI & SOCIETY,* vol. 36, no. 1, pp. 361-367, 2021.

[16]    A. F. Dugas *et al.*, "Influenza forecasting with Google flu trends," *PloS one,* vol. 8, no. 2, p. e56176, 2013.

[17]    D. Lazer, R. Kennedy, G. King, and A. Vespignani, "The Parable of Google Flu: Traps in Big Data Analysis," *Science,* vol. 343, no. 6176, pp. 1203-1205, 2014, doi: doi:10.1126/science.1248506.

[18]    A. Lazaridou *et al.*, "Mind the gap: Assessing temporal generalization in neural language models," *Advances in Neural Information Processing Systems,* vol. 34, pp. 29348-29363, 2021.

[19]    S. Amba Hombaiah, T. Chen, M. Zhang, M. Bendersky, and M. Najork, "Dynamic language models for continuously evolving content," in *Proceedings of the 27th ACM SIGKDD Conference on Knowledge Discovery & Data Mining*, 2021, pp. 2514-2524.

[20]    W. H. Organization, "Infodemic management: an overview of infodemic management during COVID-19, January 2020–May 2021," 2021.

[21]	Y. Li *et al.*, "Multi-Source Domain Adaptation with Weak Supervision for Early Fake News Detection," 2021: IEEE, doi: 10.1109/bigdata52589.2021.9671592. [Online]. Available: https://dx.doi.org/10.1109/bigdata52589.2021.9671592
[22]	J. Gramlich, "Two Years Into the Pandemic, Americans Inch Closer to a New Normal," 2022.
[23]	S. Barbosa, D. Cosley, A. Sharma, and R. M. Cesar Jr, "Averaging gone wrong: Using time-aware analyses to better understand behavior," in *Proceedings of the 25th International Conference on World Wide Web*, 2016, pp. 829-841.
[24]	Y. Dun, K. Tu, C. Chen, C. Hou, and X. Yuan, "Kan: Knowledge-aware attention network for fake news detection," in *Proceedings of the AAAI conference on artificial intelligence*, 2021, vol. 35, no. 1, pp. 81-89.
[25]	J. Gama, I. Žliobaitė, A. Bifet, M. Pechenizkiy, and A. Bouchachia, "A survey on concept drift adaptation," *ACM computing surveys (CSUR),* vol. 46, no. 4, p. 44, 2014.
[26]	I. Žliobaitė, M. Pechenizkiy, and J. Gama, "An overview of concept drift applications," in *Big data analysis: new algorithms for a new society*: Springer, 2016, pp. 91-114.
[27]	P. M. Gonçalves Jr, S. G. de Carvalho Santos, R. S. Barros, and D. C. Vieira, "A comparative study on concept drift detectors," *Expert Systems with Applications,* vol. 41, no. 18, pp. 8144-8156, 2014.
[28]	T. S. Sethi and M. Kantardzic, "On the reliable detection of concept drift from streaming unlabeled data," *Expert Systems with Applications,* vol. 82, pp. 77-99, 2017.
[29]	J. Orij, "Self-adaptation to concept drift in web-based anomaly detection," University of Twente, 2016.
[30]	J. Yang, H. Chesbrough, and P. Hurmelinna-Laukkanen, "The rise, fall, and resurrection of IBM Watson Health," Technical Report, Haas School of Business, University of California. http …, 2019.
[31]	S. Farquhar, Y. Gal, and T. Rainforth, "On statistical bias in active learning: How and when to fix it," *ICLR,* 2021.
[32]	H. Jiang, B. Kim, M. Guan, and M. Gupta, "To trust or not to trust a classifier," *Advances in neural information processing systems,* vol. 31, 2018.
[33]	N. Aslam, I. Ullah Khan, F. S. Alotaibi, L. A. Aldaej, and A. K. Aldubaikil, "Fake Detect: A Deep Learning Ensemble Model for Fake News Detection," *Complexity,* vol. 2021, pp. 1-8, 2021, doi: 10.1155/2021/5557784.
[34]	A. Galli, E. Masciari, V. Moscato, and G. Sperlí, "A comprehensive Benchmark for fake news detection," *Journal of Intelligent Information Systems,* pp. 1-25, 2022.
[35]	R. Jaiswal, U. P. Singh, and K. P. Singh, "Fake News Detection Using BERT-VGG19 Multimodal Variational Autoencoder," 2021: IEEE, doi: 10.1109/upcon52273.2021.9667614. [Online]. Available: https://dx.doi.org/10.1109/upcon52273.2021.9667614
[36]	J. Y. Khan, M. T. I. Khondaker, S. Afroz, G. Uddin, and A. Iqbal, "A benchmark study of machine learning models for online fake news detection," *Machine Learning with Applications,* vol. 4, p. 100032, 2021, doi: 10.1016/j.mlwa.2021.100032.

[37] K. Shu *et al.*, "Leveraging multi-source weak social supervision for early detection of fake news," *arXiv preprint arXiv:2004.01732,* 2020.

[38] J. P. Wahle, N. Ashok, T. Ruas, N. Meuschke, T. Ghosal, and B. Gipp, "Testing the generalization of neural language models for COVID-19 misinformation detection," in *International Conference on Information*, 2022: Springer, pp. 381-392.

[39] B. George, B. Verschuere, E. Wayenberg, and B. L. Zaki, "A guide to benchmarking COVID-19 performance data," *Public Administration Review,* vol. 80, no. 4, pp. 696-700, 2020.

[40] A. Dionysiou and E. Athanasopoulos, "Unicode Evil: Evading NLP Systems Using Visual Similarities of Text Characters," in *Proceedings of the 14th ACM Workshop on Artificial Intelligence and Security*, 2021, pp. 1-12.

[41] R. Bhalerao, M. Al-Rubaie, A. Bhaskar, and I. Markov, "Data-Driven Mitigation of Adversarial Text Perturbation," *arXiv preprint arXiv:2202.09483,* 2022.

[42] Y. Cui, W. Che, T. Liu, B. Qin, and Z. Yang, "Pre-training with whole word masking for chinese bert," *IEEE/ACM Transactions on Audio, Speech, and Language Processing,* vol. 29, pp. 3504-3514, 2021.

[43] A. Ratner, S. H. Bach, H. Ehrenberg, J. Fries, S. Wu, and C. Ré, "Snorkel: Rapid training data creation with weak supervision," in *Proceedings of the VLDB Endowment. International Conference on Very Large Data Bases*, 2017, vol. 11, no. 3: NIH Public Access, p. 269.

[44] S. Rühling Cachay, B. Boecking, and A. Dubrawski, "End-to-end weak supervision," *Advances in Neural Information Processing Systems,* vol. 34, pp. 1845-1857, 2021.

[45] A. Suprem and C. Pu, "MiDAS: Multi-integrated Domain Adaptive Supervision for Fake News Detection," *arXiv preprint arXiv:2205.09817,* 2022.

[46] M. Gruppi, B. D. Horne, and S. Adalı, "NELA-GT-2021: A Large Multi-Labelled News Dataset for The Study of Misinformation in News Articles," *arXiv preprint arXiv:2203.05659,* 2022.

[47] K. Nakamura, S. Levy, and W. Y. Wang, "r/fakeddit: A new multimodal benchmark dataset for fine-grained fake news detection," *arXiv preprint arXiv:1911.03854,* 2019.